\documentclass[conference]{IEEEconf}
\IEEEoverridecommandlockouts
\usepackage{cite}
\usepackage{amsmath,amssymb,amsfonts}
\usepackage{algorithmic}
\usepackage{graphicx}
\usepackage{textcomp}
\def\BibTeX{{\rm B\kern-.05em{\sc i\kern-.025em b}\kern-.08em
    T\kern-.1667em\lower.7ex\hbox{E}\kern-.125emX}}
    
\title{\LARGE \bf
Classifying neuromorphic data using a deep learning framework for image classification
}

\author{Roshan Gopalakrishnan, Yansong Chua\textsuperscript{*} and Laxmi R Iyer
\thanks{This research is supported by Programmatic grant no. A1687b0033 from the Singapore government’s Research, Innovation and Enterprise 2020 plan (Advanced Manufacturing and Engineering domain)}.
\thanks{Roshan Gopalakrishnan, Yansong Chua and Laxmi R Iyer are with Institute for Infocomm Research, Agency for Science, Technology and Research (A*STAR), Singapore}%
\thanks{\textsuperscript{*} corresponding author {\tt\small chuays@i2r.a-star.edu.sg}}%
}

\begin{document}



\maketitle

\begin{abstract}
In  the  field  of  artificial  intelligence,  neuromorphic computing has been around for several decades. Deep learning has however made much recent progress such that it consistently outperforms neuromorphic learning algorithms in classification tasks in terms of accuracy.  Specifically in the field of image classification, neuromorphic computing  has  been  traditionally using  either the temporal  or  rate  code for encoding static images in datasets into spike trains. It is only till recently, that neuromorphic vision sensors are widely used by the neuromorphic research community, and provides an alternative to such encoding methods.  Since then, several neuromorphic  datasets as obtained by applying such sensors on image datasets (e.g. the neuromorphic CALTECH 101) have been introduced. These data are encoded in spike trains and hence seem ideal for benchmarking of neuromorphic learning algorithms. Specifically, we  train a  deep  learning framework used for image classification on the CALTECH 101 and a collapsed version of the neuromorphic CALTECH 101 datasets. We obtained an accuracy of $91.66\%$ and $78.01\%$ for the CALTECH 101 and neuromorphic CALTECH 101 datasets respectively. For CALTECH 101, our accuracy is close to the best reported accuracy, while for neuromorphic CALTECH 101, it outperforms the last best reported accuracy by over $10\%$. This raises the question of the suitability of such datasets as benchmarks for neuromorphic learning algorithms.

\end{abstract}


\section{Introduction}

Neuromorphic computing has been around for three decades. In the late 1980s, Carver Mead developed the concept of neuromorphic computing. He described the use of analog circuits in very-large-scale integration (VLSI) systems to mimic the neuro-biological functionalities present in the human brain \cite{Mead}. Due to the complexity in understanding the complete functionality of the human brain and sensory systems, neuromorphic computing is still an open field of exploratory research.

Meanwhile neuromorphic vision sensors \cite{Tobi} has emerged as an active field of research, along with neuromorphic computing. The sensory signals from neuromorphic sensors, which is very close to human sensory signals, could be directly used by neuromorphic computing algorithms, in particular those implemented on spiking neural network (SNN) \cite{Stromatias}. Neuromorphic datasets had been created based on popular deep learning datasets, MNIST \cite{MNIST}, CALTECH \cite{CALTECH} and CIFAR-10 \cite{CIFAR}, so that the research on neuromorphic computing and its class of learning algorithms could be done on an image classification task based on standardized neuromorphic datasets, without having access to neuromorphic vision sensors.

Deep learning has been hugely successful in many areas of machine learning, in particular, with the advent of convolutional neural networks (CNN) \cite{Alex}, it has achieved human-level performance in many image classification tasks \cite{Kaiming, Taigman}. With the exponential growth of research in deep learning, and increased applications in many fields of artificial intelligence such as autonomous vehicles \cite{Hadsell, Huval} and face recognition \cite{Taigman}, deep learning is certain to drive the new industrial revolution 4.0 \cite{Tech} in the coming years.


Part of the ongoing research in deep learning is directed towards the development of deep learning accelerators. This is mainly due to the fact that firstly, deep learning networks typically need to be trained with large amount of data, and secondly the training needs to be done iteratively, which requires much computing resources for reasonably sized networks. GPUs have shown a better performance compared to CPUs. However, deep learning accelerators like TPU \cite{TPU} and other application specific integrated circuits (ASICs) are also in the competition. In comparison, neuromorphic computing chips \cite{Hasler-Marr} have been around for several decades but are generally not used for deep learning purposes, partly due to poor classification performance but also a lack of neuromorphic datasets. The lack of neuromorphic datasets is now however partially addressed when neuromorphic sensors such as the digital vision sensors (DVS) camera \cite{Lichtsteiner} are used to generate neuromorphic datasets which can be used to gauge classification performance of neuromorphic learning algorithms.


In this paper, we explored the possibility of using neuromorphic datasets in a deep learning framework. Specifically, we collapsed spike-timing data of a neuromorphic dataset into an image, which is then used to train a CNN. In doing so, we are able to obtain a classification accuracy of $78.01\%$ for the neuromorphic CALTECH dataset. While this is still some gap from the actual best classfication accuracy obtained for the CALTECH dataset ($91.66\%$) as shown in table \ref{result_tab1e}, it is already the best accuracy for neuromorphic dataset (the best reported so far being $64.2\%$ \cite{Hats}), to the best of our knowledge. The rest of the paper is organized as follows. Section \ref{M&M} elaborates on the processing of the dataset and the deep learning model used for our experiments. Section \ref{EF} further describes the experiments conducted while Section \ref{res} reports on the results obtained. The paper is then concluded in Section \ref{conc}.


\section{Materials and Methods}
\label{M&M}
\subsection{Datasets}
\label{datasets}

The deep learning models used in our experiments are trained on three datasets, namely, IMAGENET \cite{Imagenet}, CALTECH101 and neuromorphic CALTECH101 or N-CALTECH \cite{NCALTECH}.


IMAGENET: The popular image dataset comprises of 1000 categories, with over a million training images and 50,000 validation images. We use the original images in RGB format for training a deep learning model, and also a grayscale version of the original images for training another similar deep learning model. The original dataset is in JPEG format with size of $224\times224$ in each of the 3 RGB channels. In the grayscale format however, each image is converted to single channel using the formula $0.21\times R + 0.72\times G + 0.07 \times B$. However, for training the deep learning model, each grayscale image is read as three identical images, so as to maintain the same amount of trainable weights as per the deep learning model used for RGB images.


CALTECH: The original CALTECH dataset consists of images of 101 categories. Each category contains 40 to 800 image samples. Each image is roughly $300 \times 200$ in size. The original images are in RGB format, with 3 color channels. We have also the grayscale version, converted from the original dataset, also in JPEG format. Similar to IMAGENET, we duplicate the grayscale images to 3 channels when fed into the deep learning models.

N-CALTECH: We explored using a time-collapsed version of the N-CALTECH dataset. The original dataset is collapsed along the time-dimension to form static images, with pixel intensity proportional to the number of spikes at each pixel. The conversion from address event representation (AER) signals to static images is done as follows. Each pattern $p$ can be represented as a set of spike trains, one for each pixel. The spike train for pattern $p$, pixel $x$ is $s^{x, p} = \{t^{x, p}_1, t^{x, p}_2, ... t^{x, p}_n \}$ whereby each element denotes the time of spike. Here, we consider two collapsed versions of the N-CALTECH dataset, one based on only 1 saccade (N-CALTECH 1 saccade), and the other based on 3 saccades (N-CALTECH 3 saccade), both with ON polarity.


\begin{equation}
\label{eq1}
C^{x, p} = \frac{\sum_i^n t^{x, p}_i}{\max_y \sum_i^n t^{y, p}_i}
\end{equation}  

So $C^{x, p}$ is normalized by the highest spike count per pixel in pattern $p$. Note that spike counts are normalized per pattern, such that the normalizing terms differ between patterns of low and high spike counts.

\begin{figure*}[thpb]	\centering
    \begin{minipage}[b]{0.45\textwidth}
        \centering
		\includegraphics[scale=0.6]{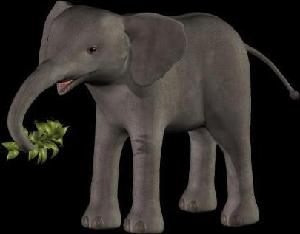} \\ (a)
     \end{minipage}
        \begin{minipage}[b]{0.45\textwidth}
        \centering
		\includegraphics[scale=0.6]{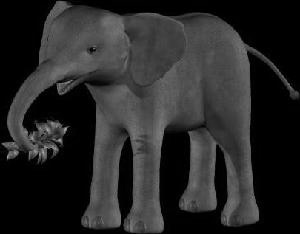} \\ (b)
    \end{minipage}
    \begin{minipage}[b]{0.45\textwidth}
        \centering
		\includegraphics[scale=0.8]{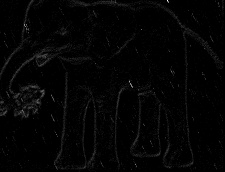} \\ (c)
     \end{minipage}
        \begin{minipage}[b]{0.45\textwidth}
        \centering
		\includegraphics[scale=0.8]{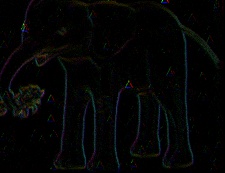} \\ (d)
    \end{minipage}
        \caption{Sample images of elephant dataset for both original CALTECH and N-CALTECH. a) Original CALTECH in 'RGB' b) Original CALTECH in grayscale c) N-CALTECH with 1 saccade and d) N-CALTECH with 3 saccade.}\label{sample_images}
\end{figure*}

For training and testing purposes, samples in each category of the entire dataset is randomly divided into training and validation images with the proportion of 85:15. The collapsed images are converted to bitmap format (.bmp) for both N-CALTECH 1 saccade and N-CALTECH 3 saccade dataset. One sample image from each CALTECH dataset is shown in figure \ref{sample_images}.

\subsection{Deep Learning Architecture}
\label{DLA}

The deep learning architecture used for the experiments is VGG-16 \cite{VGG-16}. It has a total of 16 layers, with 13 convolutional layers and 3 fully connected layers.

\begin{figure}[thpb]	\centering
    	\includegraphics[width=5cm,height=15cm]{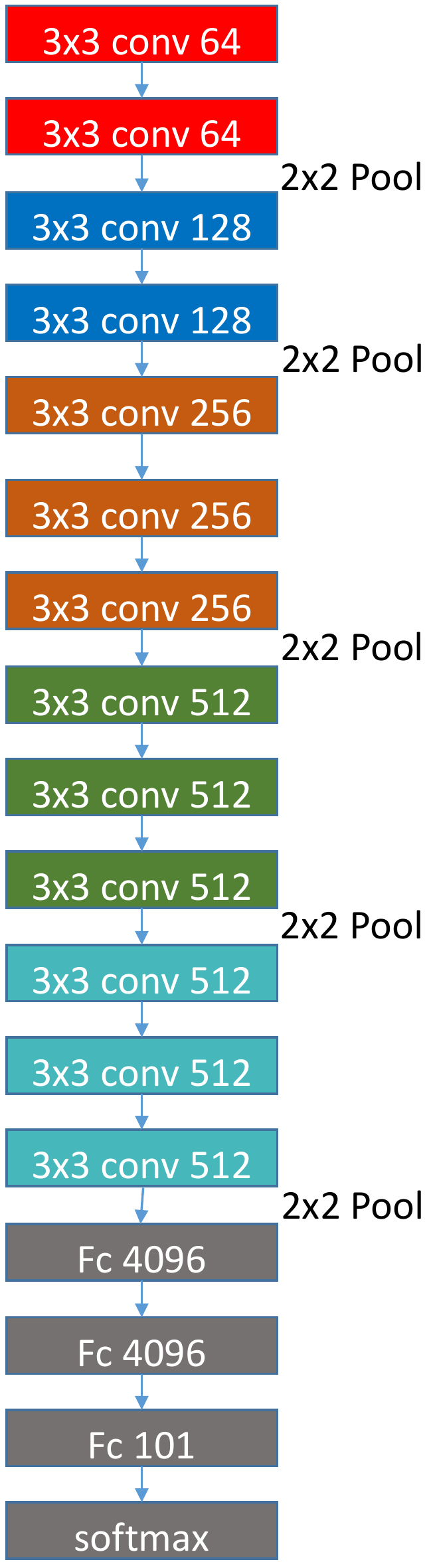} \\ (a)
        \caption{VGG-16 architecture.}\label{VGG_image}
\end{figure}

The VGG-16 architecture is shown in figure \ref{VGG_image}. It contains 5 blocks of convolutional layers (each blocks are shown in different colors). First 2 blocks contains 2 convolutional layers stacked together, followed by a max-pooling layer. The next 3 block contains 3 convolutional layers stacked together, followed by a max-pooling layer. After these 5 blocks of convolutional layers, there are 3 fully connected layers followed by a softmax layer for classification. Each computational layer contains a nonlinear activation, `RELU'. Batch normalization is used before each nonlinear activation function. A dropout of 0.5 is used in between the fully connected layers. 




\section{Experimental Framework}
\label{EF}
Entire experiments are run on a system with 4 GPUs. The deep learning models are trained on either Tensorpack \cite{Tensorpack}, or Keras \cite{Keras}, with Tensorflow backend \cite{Tensorflow}. There are two main class of experiments: one is to train the VGG-16 architecture from scratch using CALTECH-based datasets, and another is to do transfer learning: by retraining an IMAGENET (both RGB and grayscale) trained VGG-16 using CALTECH-based datasets. The loss function used is cross entropy and the Adam optimizer is applied. The weights are initialized as per \cite{Kaiming}.


\subsection{Training from scratch}
We trained the CALTECH-based datasets (CALTECH RGB, CALTECH grayscale, N-CALTECH 1 saccade and N-CALTECH 3 saccade) from scratch on the VGG-16 architecture, as described in subsection \ref{DLA}. The inputs across all datasets are kept to 224x224x3.


Both CALTECH RGB and grayscale are trained from scratch using Tensorpack, while N-CALTECH datasets are trained from scratch using Keras. All images are resized to $224 \times 224$. Each N-CALTECH image pattern $p$ is a $224 \times 224$ image with intensity value $C^{x, p}$ (eq.\ref{eq1}) at each pixel before re-sizing.

\subsection{Transfer learning}
The re-training of top layers or fully connected layers of an IMAGENET pre-trained VGG-16 using CALTECH-based datasets is described here. This approach is taken as CALTECH datasets are small and prone to over-fitting when trained from scratch, while using the features in the convolutional layers of the pre-trained VGG-16 may help to improve classification accuracy for the CALTECH-based datasets. The convolution layers of the base model is not trainable i.e their weights are freezed; whereas the fully connected layers are re-trained. We may further fine-tune the weights in the top convolutional layers in attempt to further improve classification accuracy, but this was not done in the present study.


For the CALTECH datasets, retraining of the CALTECH RGB dataset is done on the pre-trained VGG-16 using RGB IMAGENET or original IMAGENET dataset, whereas for the CALTECH grayscale dataset, retraining is done on the pre-trained VGG-16 using grayscale IMAGENET dataset.

For the case of the N-CALTECH datasets (both 1 and 3 saccades), re-training is done for each on both pre-trained VGG-16 using RGB and grayscale IMAGENET.


\section{Results}
\label{res}

The results for ``training from scratch" and ``retraining" experiments as mentioned in section \ref{EF} is shown in table \ref{result_tab1e}. Table \ref{result_tab1e} has 4 datasets in one column and two columns for training and testing accuracies for each experiment, ``training from scratch" and ``retraining". We have obtained the best accuracy of $78.01\%$ for image classification on neuromorphic CALTECH dataset, N-CALTECH 3 saccade, by retraining on the VGG-16 deep learning architecture pre-trained using grayscale IMAGENET dataset. Also, we have obtained close to best accuracy, $91.66\%$ for image classification on CALTECH dataset, CALTECH RGB, by retraining on the VGG-16 deep learning architecture pre-trained using RGB IMAGENET dataset \cite{Caltech_He}.

\begin{table}[ht]
\caption{Classification accuracies of CALTECH and N-CALTECH datasets}
\begin{center}
\begin{tabular}{|c|c|c|c|c|}
\hline
\textbf{Datasets}&\multicolumn{2}{|c|}{\textbf{Training from}}&\multicolumn{2}{|c|}{\textbf{Retraining}} \\
&\multicolumn{2}{|c|}{\textbf{scratch accuracies}}&\multicolumn{2}{|c|}{\textbf{accuracies}} \\
\cline{2-5} 
 &\textbf{\textit{Training}}& \textbf{\textit{Testing}}&\textbf{\textit{Training}}& \textbf{\textit{Testing}} \\
\hline
CALTECH & 78.15 & 76.04 & 95.18 & 91.66 \\
 RGB &  &  &  &  \\
\hline
CALTECH  & 82.44 & 78.39 & 94.43 & 91.43 \\
grayscale &  &  &  &  \\
\hline
N-CALTECH & 86.92$^{\mathrm{a}}$ & 59.97$^{\mathrm{a}}$ & 75.45 & 73.77  \\
1 saccade &  &  &  &  \\
 &  &  & (rgb) & (rgb) \\
 &  &  & 76.53 & 74.53 \\
 &  &  & (grayscale) & (grayscale) \\
\hline
N-CALTECH & 88.64$^{\mathrm{a}}$ & 57.43$^{\mathrm{a}}$ & 77.5 & 76.57 \\
3 saccade &  &  &  &  \\
&  &  & (rgb) & (rgb) \\
 &  &  & 81.01 & \textbf{\emph{78.01}} \\
 &  &  & (grayscale) & (grayscale) \\
\hline
\multicolumn{4}{l}{$^{\mathrm{a}}$Run using keras, rest using tensorpack.}
\end{tabular}
\label{result_tab1e}
\end{center}
\end{table}

\emph{Observations:} From table \ref{result_tab1e}, it is clear that the validation accuracy for retraining is better than that of training from scratch for all datasets. The training and testing accuracies in the case of ``training from scratch" are showing comparatively larger differences in comparison to those of the retrained case.

\begin{figure*}[thpb]	\centering
    \begin{minipage}[b]{0.45\textwidth}
        \centering
		\includegraphics[scale=0.45]{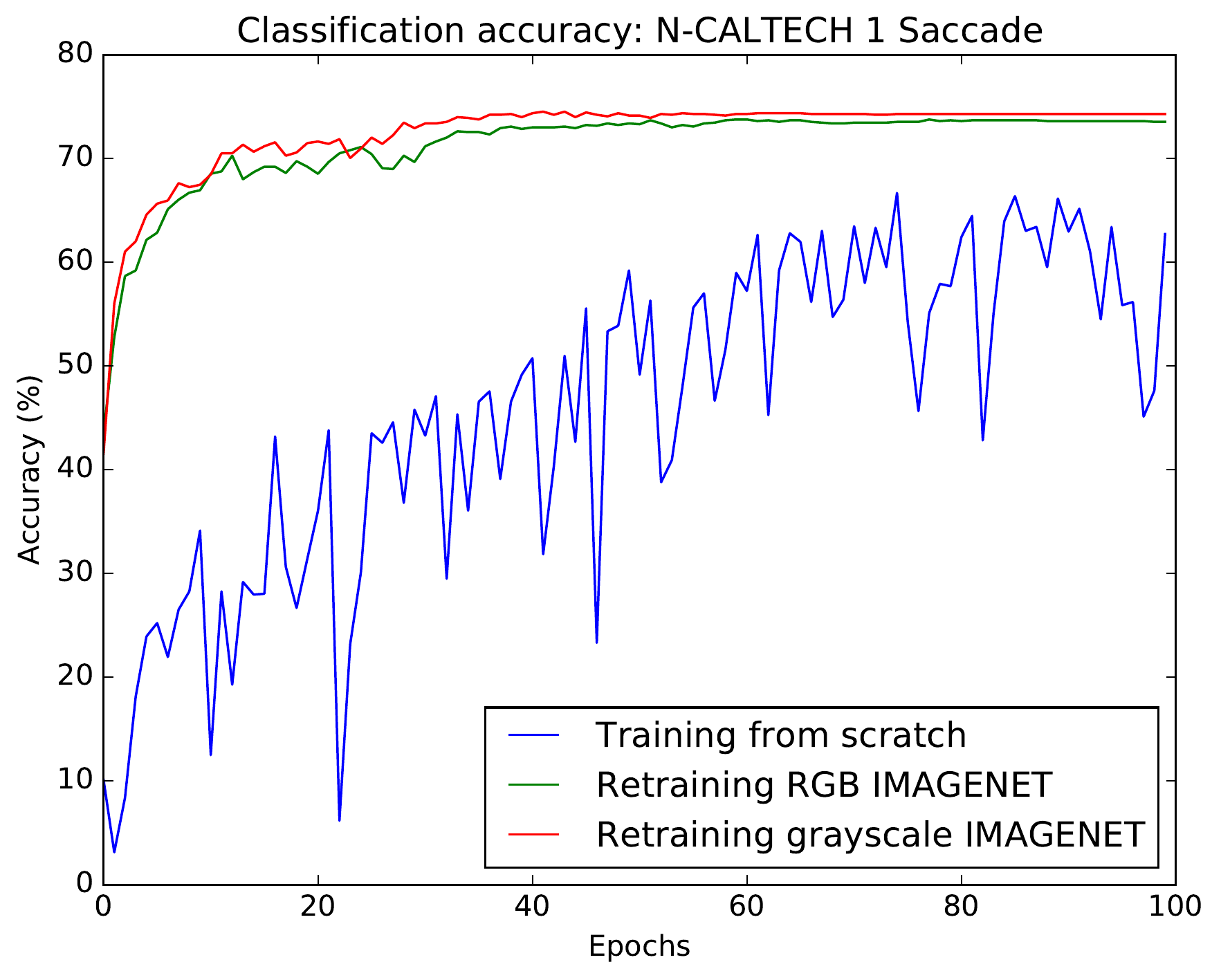} \\ (a)
     \end{minipage}
        \begin{minipage}[b]{0.45\textwidth}
        \centering
		\includegraphics[scale=0.45]{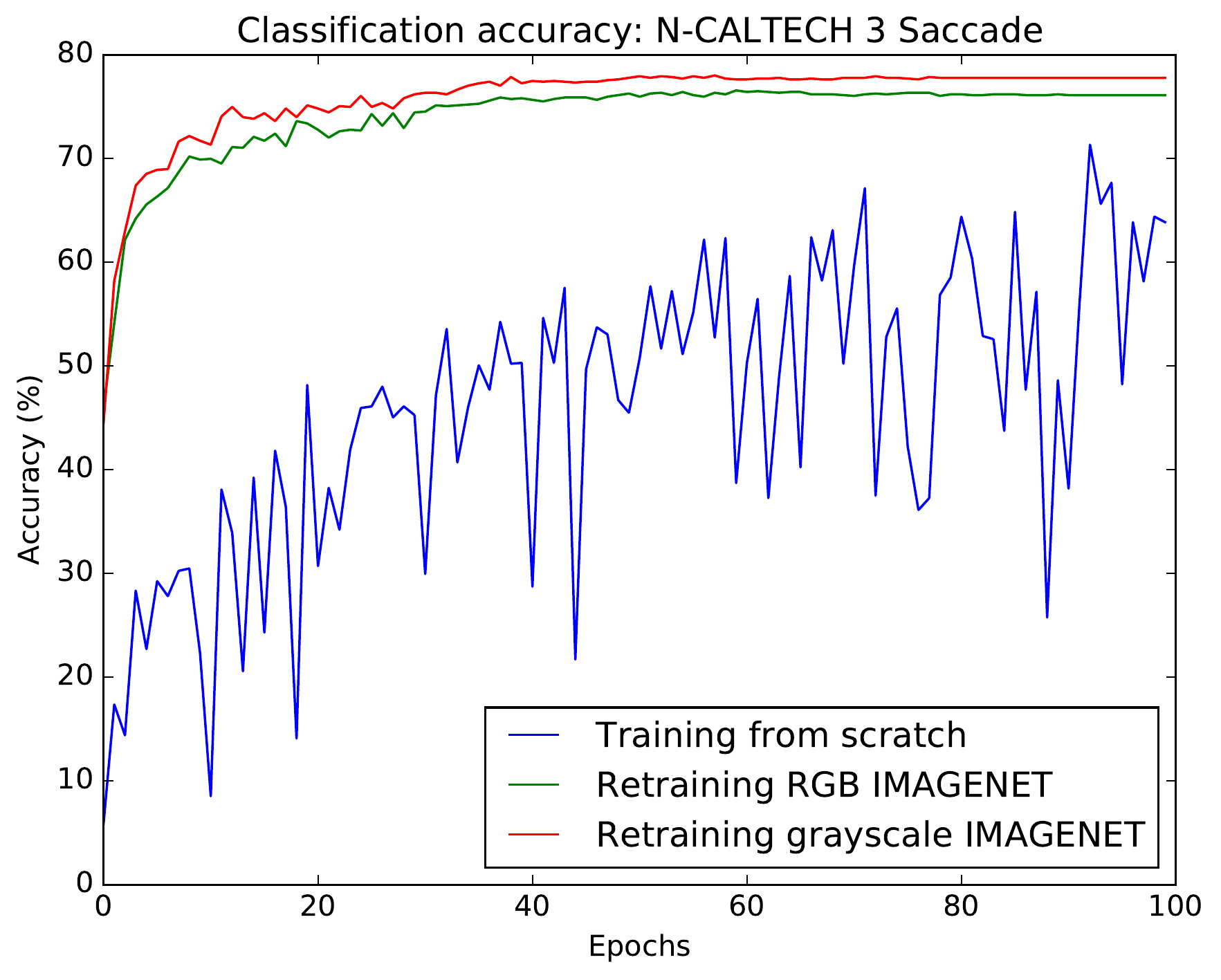} \\ (b)
    \end{minipage}
        \caption{Classification accuracy for N-CALTECH datasets to show the over-fitting problem. a) N-CALTECH with 1 saccade and b) N-CALTECH with 3 saccade.}\label{reason_image}
\end{figure*}

  
\emph{Reason:} These observations are mainly because of the lesser number of samples in the dataset. The training from scratch experiment suffers from the over-fitting problem, whereas in the case of re-training using IMAGENET pre-trained VGG-16, training is more stable and shows consistent improvement in validation accuracies, as illustrated in Fig. \ref{reason_image}. It is therefore not surprising that the pre-trained models give the best accuracies overall.

\section{Conclusion}
\label{conc}

To the best of our knowledge, we have got the best accuracy of $78.01\%$ for image classification on the neuromorphic CALTECH dataset by re-training on an IMAGENET trained VGG-16 architecture, while the best classification accuracy to date is only $64.2\%$ \cite{Hats}. This has several implications for advancing the research of learning algorithms for the neuromorphic community. Firstly, what constitutes a truly neuromorphic dataset? Our study seems to suggest that the spatial data alone in neuromorphic datasets derived from saccades across images may be sufficient for good classification results. If this is indeed the case, then using such datasets for the study of learning algorithms on SNN may not be the best way to elucidate the strengths and weaknesses of these algorithms. Another observation is that the CALTECH dataset in itself may not be sufficient for training from scratch a deep neural network for classification of its own images. This would make evaluation of learning algorithms trained on the CALTECH dataset difficult. Secondly, there is much the learning algorithms for SNN may borrow from deep learning in general. For instance, the initialization of the weights, batch normalization, are techniques that can be even more critical for SNN, given that the activations of the neurons in SNN are binary and sparse; no learning can take place if there are no activation. Thirdly, we have demonstrated how a deep learning architecture can outperform other SNN algorithms on an image classification task based on a neurmorphic dataset. This is somewhat counter-intuitive. But this also demonstrates the potential of combining neuromorphic sensors with deep learning architectures, so that we can take advantage of the event-driven nature of such sensors, and feeding the output data into a deep learning architecture for accurate classification. With the advancement of neuromorphic hardware and learning algorithms, such that reasonable classification results may be obtained using low precision weights, activation etc whilst being highly power efficient, this can be an attractive and viable proposition.


\bibliographystyle{IEEEtran}
\bibliography{reflist}

\end{document}